\documentclass[sn-aps, Numbered, twocolumn]{sn-jnl}

\usepackage{lmodern}
\usepackage{graphicx}
\usepackage{multirow}
\usepackage{amsmath,amssymb,amsfonts}
\usepackage{bm,bbm}

\DeclareMathOperator*{\argmin}{arg\,min}

\begin{document}

\title[PINNverse]{PINNverse: Accurate parameter estimation in differential equations from noisy data with constrained physics-informed neural networks}

\author*[1,2]{\fnm{Marius} \sur{Almanst\"otter}}\email{marius.almanstoetter@bsse.ethz.ch}

\author[1,2]{\fnm{Roman} \sur{Vetter}}\email{vetterro@ethz.ch}

\author[1,2]{\fnm{Dagmar} \sur{Iber}}\email{dagmar.iber@bsse.ethz.ch}

\affil*[1]{\orgdiv{Department of Biosystems Science and Engineering}, \orgname{ETH Zürich}, \orgaddress{\street{Schanzenstrasse 44}, \city{Basel}, \postcode{4056}, \state{} \country{Switzerland}}}

\affil[2]{\orgname{Swiss Institute of Bioinformatics}, \orgaddress{\street{Schanzenstrasse 44}, \city{Basel}, \postcode{4056}, \state{}\country{Switzerland}}}

\abstract{Parameter estimation for differential equations from measured data is an inverse problem prevalent across quantitative sciences. Physics-Informed Neural Networks (PINNs) have emerged as effective tools for solving such problems, especially with sparse measurements and incomplete system information. However, PINNs face convergence issues, stability problems, overfitting, and complex loss function design. Here we introduce PINNverse, a training paradigm that addresses these limitations by reformulating the learning process as a constrained differential optimization problem. This approach achieves a dynamic balance between data loss and differential equation residual loss during training while preventing overfitting. PINNverse combines the advantages of PINNs with the Modified Differential Method of Multipliers to enable convergence on any point on the Pareto front. We demonstrate robust and accurate parameter estimation from noisy data in four classical ODE and PDE models from physics and biology. Our method enables accurate parameter inference also when the forward problem is expensive to solve.}


\maketitle

\section*{Introduction}
Accurate modeling of complex phenomena in science and engineering often necessitates solving differential equations whose parameters characterize essential physical properties. Directly measuring these parameters can be difficult or impractical, which has prompted the development of inverse methods designed to infer unknown parameters from observational data.

Traditionally, inverse problems have been tackled using two main methodological frameworks. The frequentist approach seeks parameter estimates by maximizing the likelihood \cite{fisher1922, aldrich1997} of observed data given the model predictions \cite{myung2003, raue2009, frohlich2017}. However, likelihood optimization is complicated by loss landscapes that exhibit multiple modes and spurious local minima, resulting in solutions that strongly depend on initial guesses \cite{tu2002}. Consequently, multiple local minimizations from different starting points or global optimization strategies are typically employed, considerably increasing computational costs \cite{villaverde2019}. In contrast, Bayesian methods consider parameters as random variables \cite{jeffreys1939}, using Markov chain Monte Carlo algorithms to estimate posterior distributions, which naturally quantify parameter uncertainties \cite{stuart2010, luengo2020}. However, such approaches require extensive forward model evaluations, rely heavily on priors, and can face convergence challenges \cite{fang2018, cowles1996}.

Conventionally, forward problems that are formulated as differential equations (DEs) are solved using numerical schemes such as Runge--Kutta, Finite Differences, Finite Volumes, or Finite Elements. Recently, Physics-Informed Neural Networks (PINNs) \cite{raissi2019physics, karniadakis2021physics} have emerged as an attractive alternative. PINNs leverage deep neural networks in conjunction with automatic differentiation and gradient-based optimization to approximate the solutions of DEs. By incorporating DE constraints directly into the training objectives---including governing equations, initial and boundary conditions---PINNs learn to approximate the underlying physical laws. This strategy is unsupervised, mesh-free, and has successfully addressed a variety of forward and inverse problems as an alternative to classical numerical methods \cite{jamili2021parameter, mao2020, cai2022, cai2021, cuomo2022scientific, he2020, garay2024, tartakovsky2018, wei2023}. Nevertheless, employing PINNs can still present challenges, such as complex loss landscapes due to soft enforcement of constraints and difficulties encountered in scenarios involving shock waves \cite{fuks2020, krishnapriyan2021}. Consequently, several enhancements have been proposed over recent years to enforce physical laws more effectively and to expand the applicability of PINNs. They include respecting temporal causality \cite{wang2022}, including Fourier features in the first layer \cite{tancik2020}, curriculum-based training \cite{krishnapriyan2021}, adaptive resampling strategies during training \cite{wu2022, daw2023}, adaptive weights during training \cite{xiang2022, wang2022ntk, mcclenny2023} and augmenting the training process with additional equations \cite{yu2021}.

When training PINNs, a data loss term is introduced that competes with the physics-based loss component. This interplay has recently prompted increased interest in multi-objective optimization techniques and the exploration of Pareto fronts within the context of PINNs \cite{rohrhofer2021, heldmann2023, bischof2021, lazovskaya2023, lu2023, wong2025}. These studies typically emphasize on adaptive weighting schemes to manage the relative contributions of various loss terms, or apply evolutionary algorithms, such as NSGA-II \cite{calyanmoy2002}, explicitly targeting multi-objective optimization. However, these evolutionary methods often involve substantially greater computational expense compared to conventional PINN training. The adaptive weighting schemes introduce additional, often sensitive, hyperparameters that themselves require careful tuning, adding another layer of complexity to the problem.

In inverse problem scenarios with substantial noise in the observational data, simultaneously minimizing data and physics losses to zero is unfeasible, necessitating a deliberate trade-off between these competing objectives. Furthermore, equal balancing of losses is typically undesirable, as it can result in overfitting the data noise instead of accurately capturing the underlying physics.

To overcome these limitations, we propose a constrained optimization framework tailored to identify optimal PINN solutions that accurately fit observational data while strictly enforcing physical constraints. Previous investigations into constrained optimization approaches within PINNs have demonstrated networks structured to exactly fulfill initial and boundary conditions, while employing augmented Lagrangian methods to minimize physics and data losses \cite{lu2021}. Additionally, stochastic augmented Lagrangian methods have been effectively applied to train PINNs \cite{dener2020}. However, the augmented Lagrangian requires nested loops for training, which are typically not employed in deep learning, and increases computational complexity. To overcome this limitation, we propose employing the Modified Differential Method of Multipliers (MDMM) \cite{platt1987}, an optimization strategy that simultaneously updates the neural network parameters, differential equation parameters, and associated Lagrange multipliers in parallel. This method is fully compatible with modern state-of-the-art optimizers, such as Adam \cite{kingma2017} or Adan \cite{xie2024}, and importantly, it does not incur additional computational costs beyond those of standard PINN training methods. We demonstrate on four classical examples that the proposed training paradigm outperforms standard PINNs, especially with high levels of noise in the data, and traditional optimizers such as the Nelder--Mead \cite{nelder1965} when the initial parameter guess is inaccurate.

\section*{Results}

\begin{figure*}
    \centering
    \includegraphics[width=\textwidth]{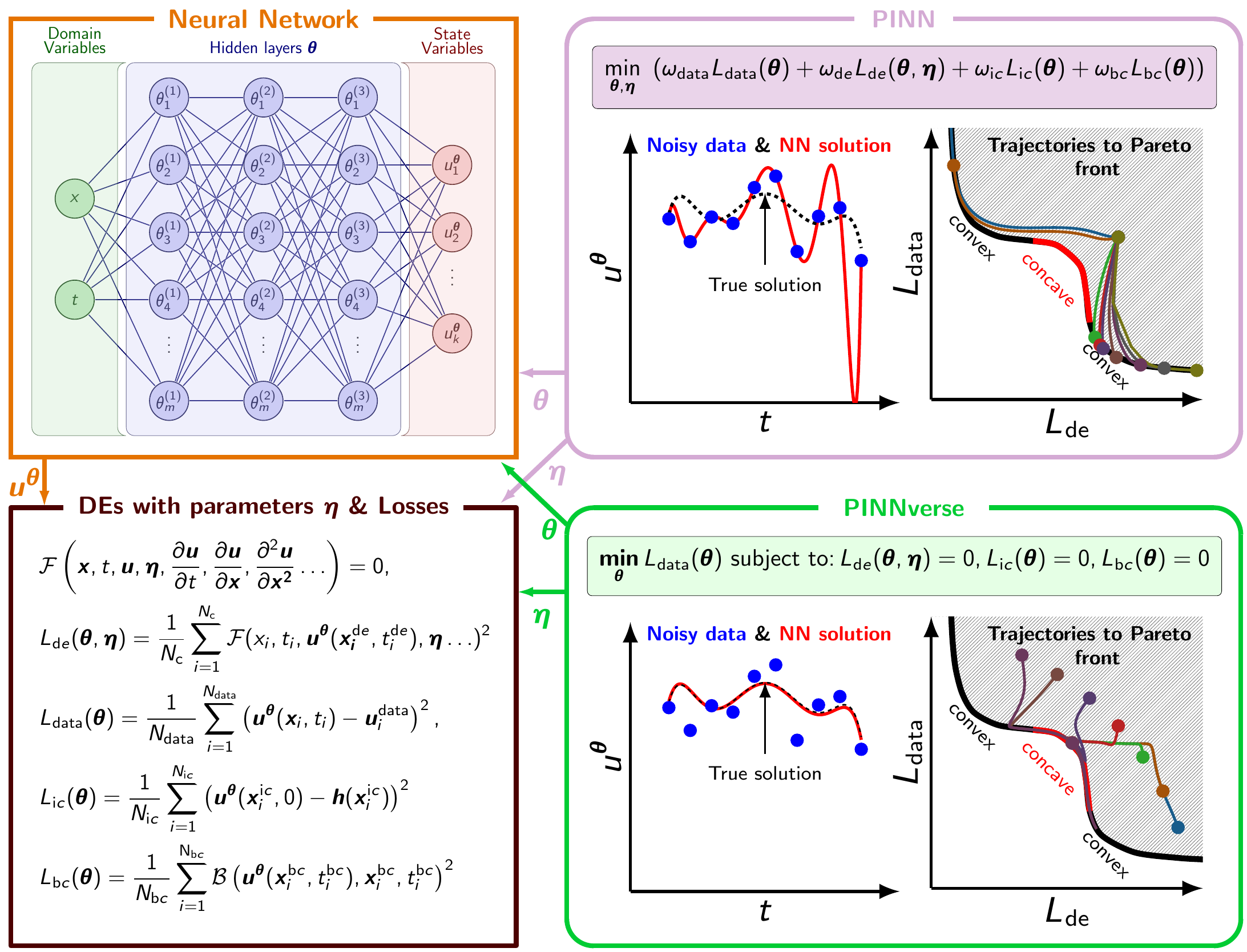}
    \caption{\textbf{Schematic representation of the difference between PINN and PINNverse.} When approximating solutions to differential equations (DEs) in residual form $\mathcal{F}=0$ with a neural network (NN), the architecture utilizes hidden layers (parameter set $\bm{\theta}$) to map input variables including spatial coordinates $\bm{x}$ and time $t$ to the solution space of state variables, represented by a $k$-dimensional vector of functions $\bm{u}^{\bm{\theta}}$ (orange box). The training process of a PINN (purple box) minimizes a composite loss function incorporating terms penalizing deviations of the predicted solution $\bm{u}^{\bm{\theta}}$ from observed noisy data points $\bm{u}^\textrm{data}_i$ (blue dots) ($L_{\textrm{data}}$), as well as terms that penalize violations of the differential equations ($L_\mathrm{de}$) and of the initial and/or boundary conditions ($L_{\textrm{ic}}$, $L_{\textrm{bc}}$). The predicted NN solution typically suffers from overfitting and misses non-convex parts (solid red line) of the Pareto front (solid black line). Examples of trajectories starting from initial points within the feasible region (dashed area) and leading to this front are schematically visualized for a 2D subspace. With PINNverse (green box), the data, IC and BC losses are formulated as external constraints under which the optimization is carried out, which avoids overfitting and allows the trajectories to converge also on convex parts of the front.}
    \label{fig:PINNverse}
\end{figure*}

\subsection*{PINNverse restates the training paradigm}

PINNverse is a reformulation of the training paradigm for PINNs that addresses their fundamental limitations in solving inverse problems. The key innovation lies in the handling of multiple competing objectives that arise in PINNs. Traditional PINNs employ a composite loss function that linearly combines data-fitting terms with physics constraints using fixed weights (Fig.~\ref{fig:PINNverse}, purple box; Methods). This weighted-sum approach creates a multi-objective optimization problem that struggles with complex Pareto fronts (the set of solutions that minimize the loss), particularly those with non-convex regions where conventional gradient-based methods generally fail to find balanced solutions. A minimization with stochastic gradient descent algorithms lets even trajectories originating from the same initial point consistently converge towards different convex regions of the solution space depending on the used learning rate and starting point. The resulting reachable Pareto front represents only the convex portions of the solution landscape, which can lead to a strong emphasis of the NN solution on data enforcement (small data loss), and thus overfitting. 

PINNverse restructures this approach by reformulating the problem as constrained optimization rather than weighted summation (Methods). We designate the data-fitting term as the primary objective while transforming physics-based terms (differential equations, initial conditions, and boundary conditions) into explicit constraints (Fig.~\ref{fig:PINNverse}, green box). This shift from penalty-based regularization to explicit constraint enforcement substantially changes how physics information is incorporated into the learning process.

To solve this constrained problem, we employ the Modified Differential Method of Multipliers, which enables convergence to any point on the Pareto front, including those in concave regions that traditional methods would miss. The primary advantage of the MDMM lies in its inherent capability to converge towards saddle points of the loss landscape, which constitute the optimal solutions within a Lagrangian formulation \cite{sun2005}. This property enables it to efficiently identify balanced solutions that simultaneously fulfill data-fitting objectives and physics-based constraints, ensuring that neither is compromised in favor of the other. Unlike conventional approaches in which certain loss terms may dominate others during optimization, PINNverse ensures that all physics constraints are properly enforced while simultaneously fitting the available data.

\subsection*{Experimental design}

We evaluate PINNverse against PINNs and the widely adopted Nelder--Mead optimization algorithm \cite{nelder1965}, specifically using the implementation in the SciPy library \cite{virtanen2020}. Since only Nelder--Mead and PINNverse naturally support bounds, we use reasonable parameter bounds only for these two methods, and ensure parameter positivity by exponential transformation for the PINN. Our experimental framework encompasses four benchmark problems---two ordinary differential equations (ODEs) and two partial differential equations (PDEs)---to assess the generalizability, robustness and accuracy of our approach.

To isolate the effects of our methodological contribution against the PINN, we employ identical neural network architectures, optimization algorithms, learning rate schedules, and initialization procedures across both PINNverse and standard PINN implementations (Methods). In all cases we have used the Adan optimizer \cite{xie2024} for training. Our application of MDMM in PINNverse is based on a specific PyTorch implementation (\url{https://github.com/crowsonkb/mdmm}). This controlled experimental design ensures comparability and fairness: observed performance differences can be attributed solely to the modified gradient update strategy of PINNverse.

\subsection*{Kinetic reaction model}

\begin{figure*}
    \centering
    \includegraphics[width=\textwidth]{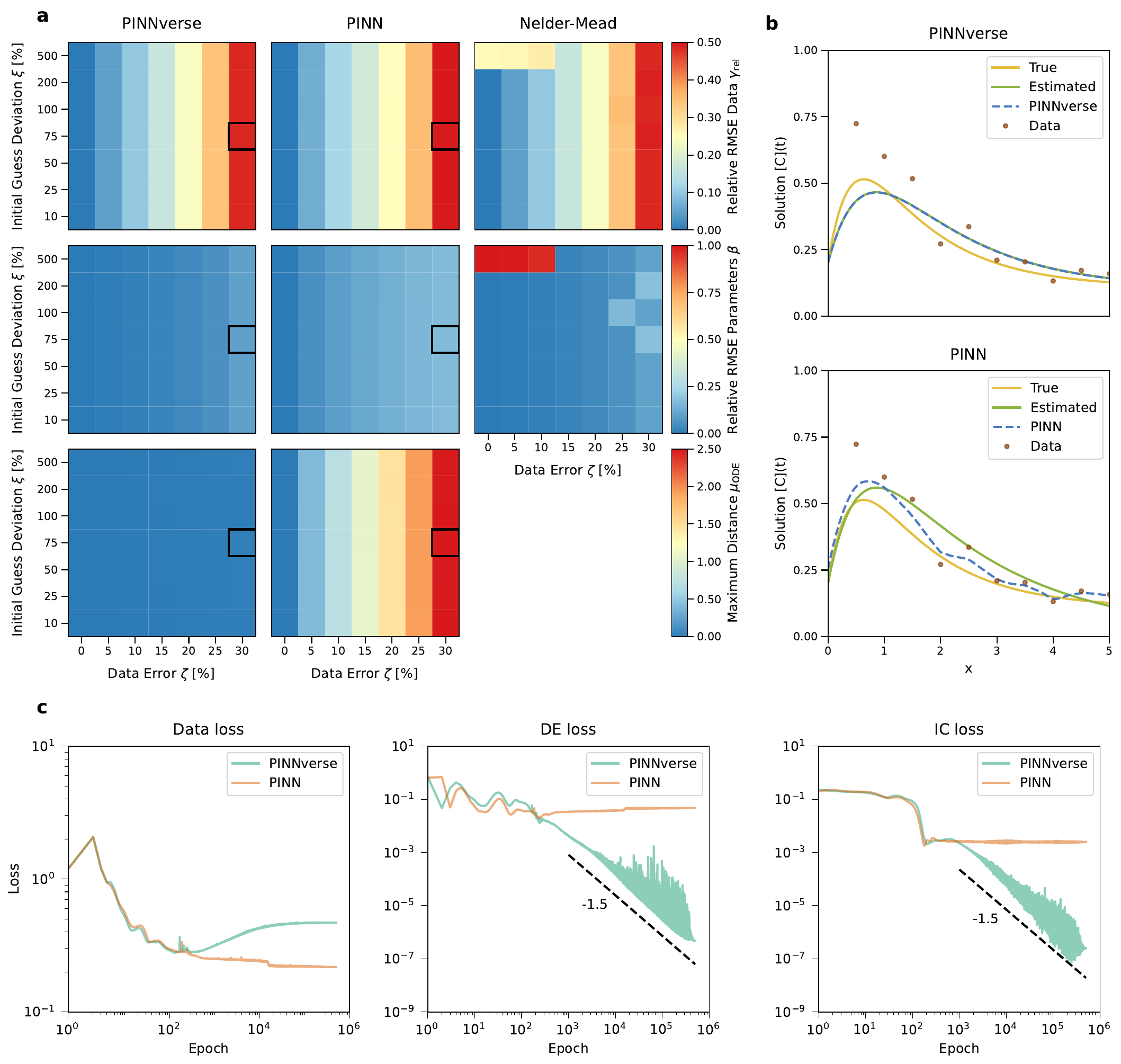}
    \caption{
    \textbf{Parameter estimation performance in the kinetic reaction ODE model.}
    \textbf{a}, Heatmaps depicting performance metrics across varying noise levels in the data, $\zeta$, and deviations in initial parameter guesses, $\xi$ (Methods). The black square highlights the scenario $\zeta=25\%$, $\xi=75\%$ analyzed in detail in subsequent panels.
    \textbf{b}, Comparison of trajectories for species $[C](t)$, generated using estimated parameters (green curve), true parameters (yellow curve), neural network predictions (blue curve) and the corresponding noisy observational data (brown dots).
    \textbf{c}, Training loss evolution for PINNverse and conventional PINN. Data, differential equation (DE) and initial condition (IC) losses are depicted. For PINNverse, a power law was fitted to the DE and IC losses after 1000 epochs (shifted dashed lines) with indicated exponents.}
    \label{fig:reaction}
\end{figure*}

We begin by examining a reduced version of a more detailed nonlinear reaction model previously employed in the context of parameter estimation using PINNs \cite{bibeau2024}, involving four distinct species, labeled $A$, $B$, $C$ and $D$:
\begin{equation*}
A \underset{k_2}{\overset{k_1}{\rightleftharpoons}} B + C, \quad C \underset{k_4}{\overset{k_3}{\rightleftharpoons}} D,
\end{equation*}
where the double arrows indicate reversible reactions at indicated rates $k_1$ to $k_4$. The system of ODEs describing the temporal evolution of the concentrations reads
\begin{align*}
    \frac{d[A]}{dt} &= -k_1 [A] + k_2 [B] [C], \\
    \frac{d[B]}{dt} &= k_1 [A] - k_2 [B] [C], \\
    \frac{d[C]}{dt} &= k_1 [A] - k_2 [B] [C] - k_3 [C] + k_4 [D], \\
    \frac{d[D]}{dt} &= k_3 [C] - k_4 [D].
\end{align*}
with initial conditions 
\begin{align*}
    &[A](0)=1.0, \quad [B](0)=0,\\
    &[C](0)=0.2, \quad [D](0)=0.
\end{align*}
ODE model systems of this type are widely used to characterize chemical reactions and dynamic interactions in complex biological systems. For instance, minimal models of glucose regulation of this form have been used to estimate insulin sensitivity by analyzing plasma glucose and insulin dynamics during glucose tolerance tests \cite{bergman1979}. They are generally popular toy models for parameter inference, e.g., in the form of the oscillatory Lotka--Volterra equations \cite{babadzanjanz2003}.

We generated synthetic data by numerically solving the system at ten different time points using known (ground-truth) kinetic parameters $\bm{\eta}_{\text{true}} = [k_1, k_2, k_3, k_4] = [1.5, 0.5, 1, 0.1]$. Parameter bounds $\bm{\eta}^{\textrm{lower}} = [0,0,0,0]$ and $\bm{\eta}^{\textrm{upper}} = [10,4,7,0.7]$ were used in PINNverse and Nelder--Mead. The PINN and PINNverse were trained for 500,000 epochs.

Figure~\ref{fig:reaction}a quantifies the performance of each method with respect to data quality and uncertainty in parameter initialization. The conventional PINNs, PINNverse, and Nelder--Mead all achieve similar relative root mean squared error (RMSE, $\gamma_\mathrm{rel}$, Methods) between model predictions and noisy observations (top row), growing as expected with increasing measurement noise ($\zeta$, Methods). However, Nelder--Mead struggles when initialized far from the true parameter values ($\xi=500\%$; Methods), getting trapped in local minima. PINNverse achieves a mean improvement factor over Nelder--Mead of approximately 370 in this scenario.

The second row of Fig.~\ref{fig:reaction}a evaluates the parameter estimation accuracy $\beta$ (Methods). While the conventional PINN yields accurate parameters exclusively under noise-free conditions and quickly deteriorates as measurement noise increases, PINNverse and Nelder--Mead generally retain higher accuracy. PINNverse achieves a 3.8-fold improvement in $\beta$ compared to standard PINNs (mean across all tested scenarios). Nonetheless, Nelder--Mead fails when initialized far from the true parameter values, and occasionally also underperforms at higher noise levels ($\zeta \geq 25\%$), where PINNverse demonstrates a 1.2-fold improvement in $\beta$.

The bottom row of Fig.~\ref{fig:reaction}a illustrates the maximum deviation $\mu_{\text{ODE}}$ (Methods) between the neural network predictions and the true solutions at the inferred parameters---a metric in which the advantages of PINNverse are particularly evident. While conventional PINNs violate physical constraints as measurement noise increases, PINNverse consistently remains conforming. On average across all noise-affected scenarios, PINNverse yields an 88-fold improvement in $\mu_{\text{ODE}}$ compared to the standard PINN.

To illustrate the overfitting tendencies of conventional PINNs and the physics-conforming behavior of PINNverse, a representative scenario ($\zeta=25\%$, $\xi=75\%$, black squares in Fig.~\ref{fig:reaction}a) is shown in Fig.~\ref{fig:reaction}b. Predictions of concentration $[C](t)$ reveal perfect alignment between PINNverse predictions (blue curve) and numerical solutions computed from the inferred parameters (green curve). Despite considerable observational noise (brown dots), PINNverse closely approximates the underlying true dynamics (yellow curve). In contrast, conventional PINN predictions deviate substantially from numerical solutions, strongly overfitting the noisy data at the expense of physical accuracy.

This pronounced bias of PINNs toward data loss minimization, neglecting physics and initial condition losses due to inherent non-convexities in the multi-objective landscape, is also apparent in convergence plots (Fig.~\ref{fig:reaction}c). In contrast, PINNverse achieves substantial simultaneous reductions in physics and initial condition losses consistent with algebraic convergence beyond 1000 epochs: $L_\mathrm{de,ic}\sim\text{epoch}^{-a}$ with exponents $a=1.5255\pm0.0009$ for DE and $a=1.5108\pm0.0005$ (s.e.) for IC.

\subsection*{FitzHugh--Nagumo model}

\begin{figure*}
    \centering
    \includegraphics[width=\textwidth]{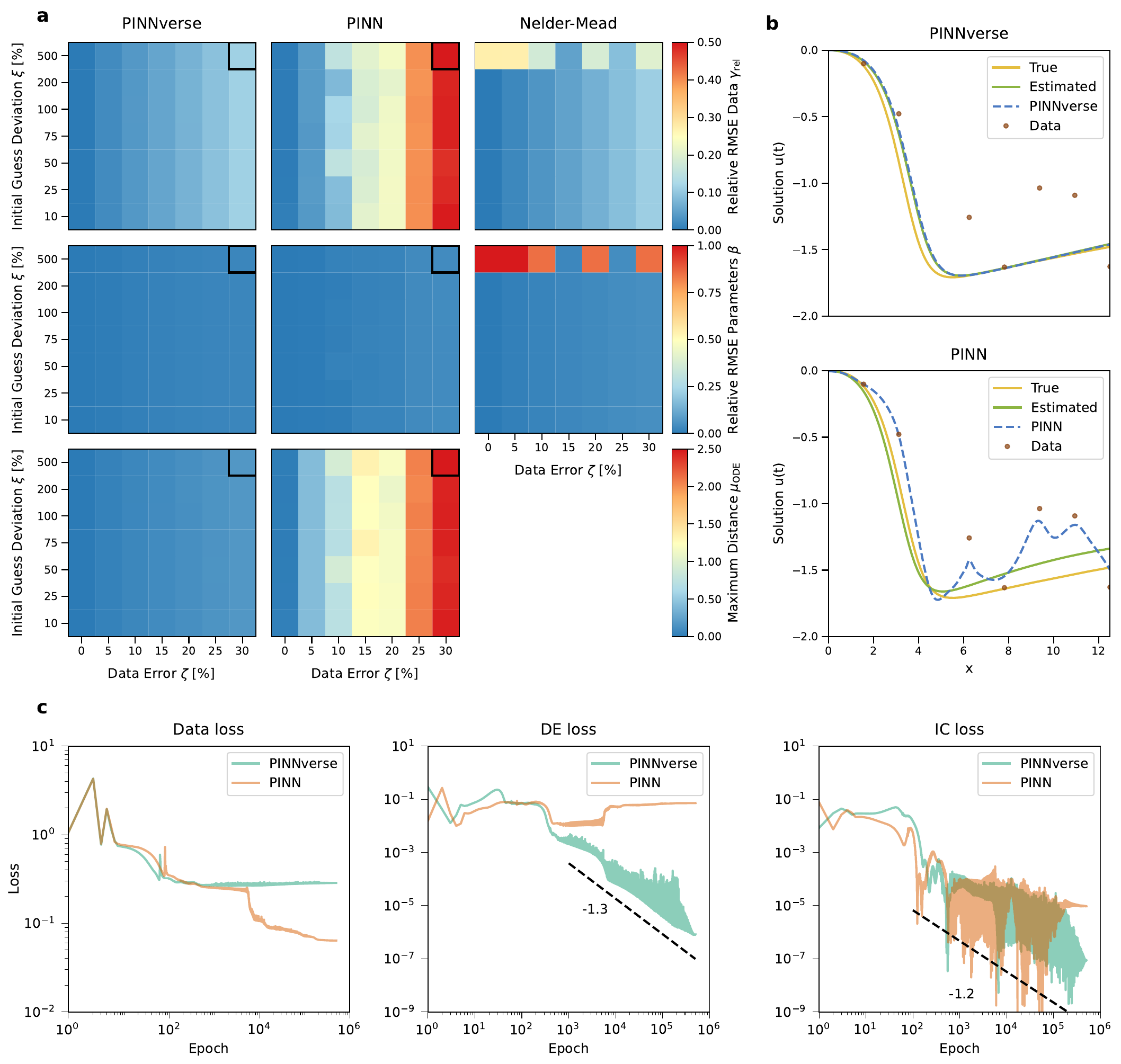}
    \caption{\textbf{Parameter estimation performance in the FitzHugh--Nagumo ODE model.}
    \textbf{a}, Heatmaps depicting performance metrics across varying noise levels in the data, $\zeta$, and deviations in initial parameter guesses, $\xi$ (Methods). The black square highlights the scenario $\zeta=25\%$, $\xi=500\%$ analyzed in detail in subsequent panels.
    \textbf{b}, Comparison of trajectories for the excitable variable $u(t)$, generated using estimated parameters (green curve), true parameters (yellow curve), neural network predictions (blue curve), and the corresponding noisy observational data (brown dots).
    \textbf{c}, Training loss evolution for PINNverse and conventional PINN. Data, differential equation (DE) and initial condition (IC) losses are depicted. For PINNverse, a power law was fitted to the DE and IC losses after 1000 epochs (shifted dashed lines) with indicated exponents.}
    \label{fig:fitz_hugh_nagumo}
\end{figure*}

Originally introduced as a simplification of the Hodgkin--Huxley model \cite{hodgkin1952}, the FitzHugh--Nagumo (FHN) equations \cite{fitzhugh1961, nagumo1962} capture essential neuronal phenomena such as excitability and refractoriness. Their mathematical tractability makes them a powerful tool for analyzing neuronal firing patterns and stochastic behavior. Accurate parameter estimation in the FHN model is essential for quantitatively interpreting and predicting neuronal responses to different stimuli. Techniques utilizing noisy voltage-clamp data have effectively recovered intrinsic neuronal parameters \cite{che2012}, while maximum likelihood methods based on spike timing have characterized neuronal responses without amplitude information \cite{doruk2019}. Recently, approximate Bayesian computation approaches with structure-preserving numerical schemes have robustly estimated parameters from stochastic neuronal data, enhancing the model's applicability \cite{samson2024}. Additionally, the FHN equations have gained prominence in physics-informed neural network literature \cite{rudi2022, bizzi2024}, which is why they are our next benchmark.

We use the classical two-dimensional spatially homogeneous form of the FitzHugh--Nagumo model:
\begin{align*}
\frac{\mathrm{d}u}{\mathrm{d}t} &= u - \frac{u^3}{3}-v,\\
\frac{\mathrm{d}v}{\mathrm{d}t} &= \frac{u+a-b v}{r},\\
\end{align*}
with initial conditions 
\begin{align*}
    u(0)&=0, \quad v(0)=0
\end{align*}
where $u(t)$ is the excitable (membrane potential-like) variable and $v(t)$ represents the slow recovery variable. Parameters $a$, $b$, and $r$ regulate threshold dynamics, recovery rates, and timescale separation, respectively.

To subject the optimization methods to a particularly tough challenge with sparse data, we use only seven data points as input, calculated from the solution for $\bm{\eta}_{\text{true}} = [a, b, r] = [0.7, 0.8, 12.5]$ plus noise (Methods). Parameter bounds $\bm{\eta}^{\textrm{lower}} = [0,0,0]$ and $\bm{\eta}^{\textrm{upper}} = [10,10,100]$ were set. The same number of training epochs was used as for the kinetic reaction model.

The observed trend is broadly consistent with the reaction model, although with several noteworthy differences (Fig.~\ref{fig:fitz_hugh_nagumo}a). The standard PINN consistently exhibits substantially higher relative data error ($\gamma_{\textrm{rel}}$) compared to PINNverse, particularly as the measurement noise level increases. Across all tested scenarios, PINNverse achieves a 4.7-fold mean improvement in $\gamma_{\textrm{rel}}$. While the Nelder--Mead algorithm performs similarly well with good initial guesses, it again fails when initialized far from the true parameters ($\xi=500\%$), where PINNverse outperforms it by a factor of approximately 221.

PINNverse also yields a substantially better parameter estimate than the PINN  (Fig.~\ref{fig:fitz_hugh_nagumo}a, middle row), with a 3.5-fold improvement in $\beta$ (mean over all tested scenarios). Under the challenging condition of an initial parameter guess deviation of $\xi=500\%$, the Nelder--Mead algorithm fails to find the right parameter also in this benchmark. Furthermore, the solutions predicted by the standard PINN deviate markedly from the numerical reference solution, whereas PINNverse predictions closely match it (Fig.~\ref{fig:fitz_hugh_nagumo}a, bottom row).

In Fig.~\ref{fig:fitz_hugh_nagumo}b we provide a representative example illustrating these observations through component $u(t)$ for the scenario with largest data noise and deviation in the initial guess (black squares in Fig.~\ref{fig:fitz_hugh_nagumo}a). The PINNverse prediction closely follows the true trajectory, while that of the PINN deviates by overfitting the noisy data. Also when the estimated parameters are used in a numerical solver, the computed solution match the true solution only for PINNverse, not for the PINN.

In convergence plots (Fig.~\ref{fig:fitz_hugh_nagumo}c), a similar picture is observed for the FHN equations as in the reaction model: The PINN systematically fails to reduce the DE and IC losses after about 1000 epochs. PINNverse, on the other hand, reduces them further, approximately following power laws $L_\mathrm{de,ic}\sim\text{epoch}^{-a}$ with slightly lower algebraic convergence rates $a=1.3340\pm0.0007$ and $a=1.159\pm0.001$ (s.e.) for DE and IC, respectively.

\subsection*{Fisher--KPP model}

\begin{figure*}
    \centering
    \includegraphics[width=\textwidth]{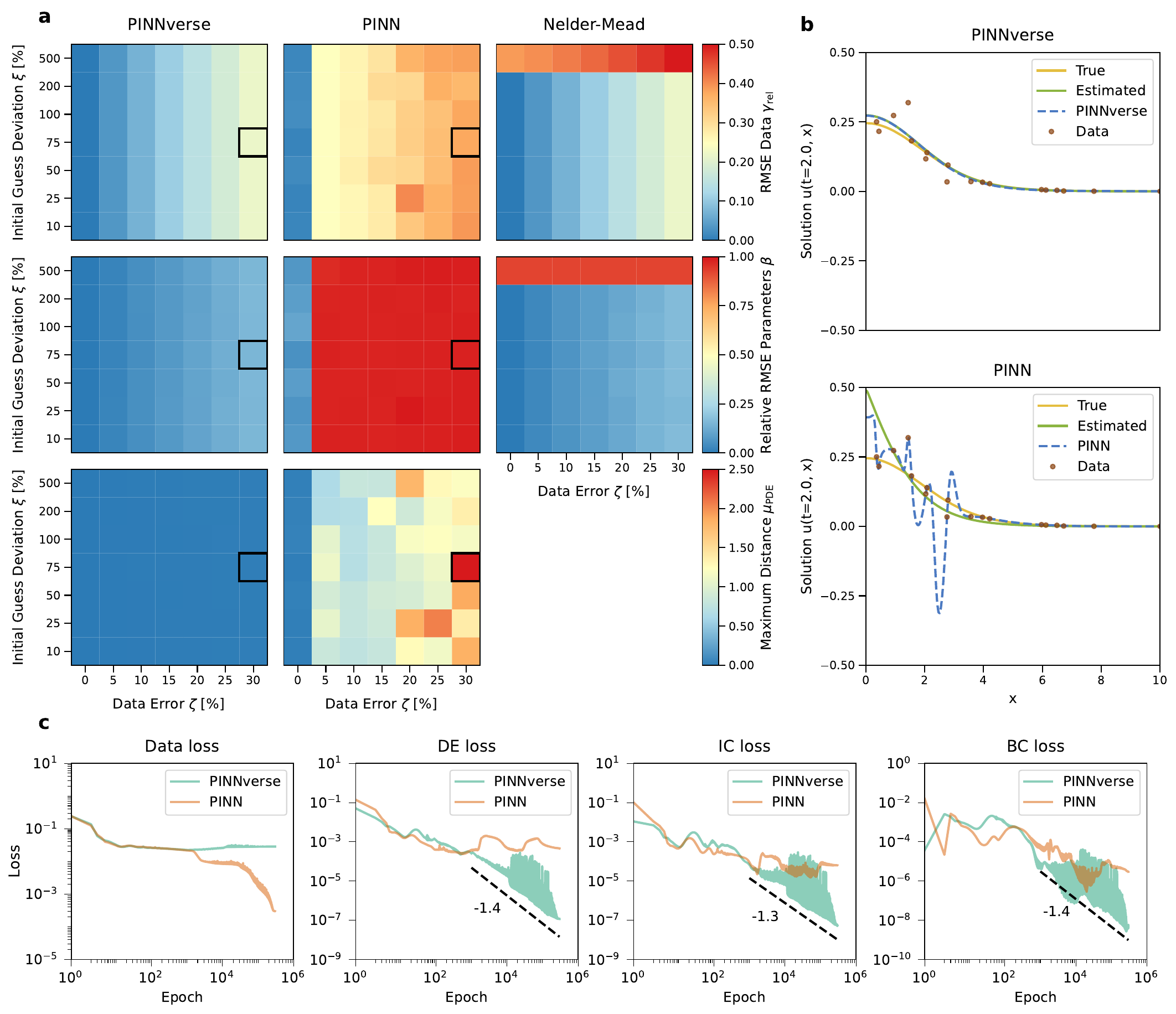}
    \caption{\textbf{Parameter estimation performance in the Fisher--KPP PDE model.}
    \textbf{a}, Heatmaps depicting performance metrics across varying noise levels in the data, $\zeta$, and deviations in initial parameter guesses, $\xi$ (Methods). The black square highlights the scenario $\zeta=25\%$, $\xi=75\%$ analyzed in detail in subsequent panels.
    \textbf{b}, Comparison of trajectories for the cell concentration $u(x)$ at time point $t=2$, generated using estimated parameters (green curve), true parameters (yellow curve), neural network predictions (blue curve), and the corresponding noisy observational data (brown dots).
    \textbf{c}, Training loss evolution for PINNverse and conventional PINN. Data, differential equation (DE), initial condition (IC) and boundary condition (BC) losses are depicted. For PINNverse, a power law was fitted to the DE, IC and BC losses after 1000 epochs (shifted dashed lines) with indicated exponents.}
    \label{fig:fisher}
\end{figure*}

The Fisher--KPP equation, pioneered by Fisher and Kolmogorov, Petrovsky and Piskunov \cite{fisher1937, kolmogorov1937}, is the first of two PDEs in our test suite. It has emerged as a fundamental mathematical framework to model spatiotemporal population dynamics across diverse biological contexts, from two-dimensional cell spreading during skeletal tissue regeneration \cite{sengers2007} to motility and proliferation in circular barrier assays relevant to wound healing \cite{simpson2013} and malignant cell proliferation influenced by TGF-$\beta$ signaling \cite{wang2009}. Recently, PINNs have also been explored for solving this equation \cite{rohrhofer2025, sultan2024}. We use its one-dimensional form with the following initial and boundary conditions:
\begin{align*}
    \frac{\partial u}{\partial t} &= D \frac{\partial^2 u}{\partial x^2} + \rho u (1 - u) \\
    u(x, 0) &= \frac{1}{10} e^{-x} \\
    \frac{\partial u(x,t)}{\partial x} &= 0 \quad \text{at} \quad x \in \{0, 10\}
\end{align*}
Here, $u(x,t)$ represents the normalized population density at position $x$ and time $t$. $D$ is the diffusion coefficient characterizing the random motility of the cells. The nonlinear source term $\rho u(1 - u)$ accounts for logistic growth, representing cell proliferation at rate $\rho$, constrained by carrying capacity limitations. The exponential initial condition represents a localized cell distribution gradually decreasing with distance. Zero-flux (Neumann) boundary conditions ensure that there is no cell transfer across the domain boundaries. A particularly challenging feature of this PDE is that it admits traveling wave solutions whose propagation velocity is determined by the underlying model parameters.

To generate the synthetic measurement data, we used ground-truth kinetic parameters $\bm{\eta}^{\text{true}} = [\mathrm{D}, \rho] = [0.5, 1]$ and perturbed the exact solution at a total of 18 points in time and space as before, at observation times $t=1$ and $t=2$. $\bm{\eta}^{\textrm{lower}} = [0.1,0.5]$ and $\bm{\eta}^{\textrm{upper}} = [0.5, 6]$ served as parameter bounds for PINNverse and Nelder--Mead. We trained the PINNverse and PINN for 300,000 epochs.

To assess the performance on the Fisher--KPP, we quantified the \emph{absolute} deviation between model prediction and data, $\gamma_{\textrm{abs}}$ (Methods), because solution values can approach zero. PINNverse is seen to consistently and substantially outperform the PINN whenever measurement noise is non-zero (Fig.~\ref{fig:fisher}a), by all metrics considered. PINNverse achieves a 12-fold mean improvement in $\gamma_{\mathrm{rel}}$ and 48-fold in $\beta$ across all tested scenarios. Here, a peculiar weakness of the normal PINN becomes apparent: While the solution learned by the PINN strongly overfits the noisy data (Fig.~\ref{fig:fisher}b), the numerical solution obtained by solving the PDE with the parameters inferred by the PINN does not approximate the data well, because the parameter estimate is poor (Fig.~\ref{fig:fisher}a, middle column). PINNverse, on the other hand, robustly finds the optimal parameters and reasonably approximates the data (Fig.~\ref{fig:fisher}a, left column).

In convergence plots (Fig.~\ref{fig:fisher}c), we observe that PINNverse manages to reduce all physical loss terms (DE, IC and BC), while the classical PINN starts overemphasizing the data loss beyond about 1000 epochs at the expense of struggling with the physical losses. With PINNverse, approximate algebraic convergence in the number of epochs is observed: $L_\mathrm{de,ic,bc}\sim\text{epoch}^{-a}$ with $a=1.422\pm0.001$, $a=1.265\pm0.002$ and $a=1.413\pm0.002$ (s.e.) for DE, IC and BC, respectively.

\subsection*{Burgers' equation}

\begin{figure*}
    \centering
    \includegraphics[width=\textwidth]{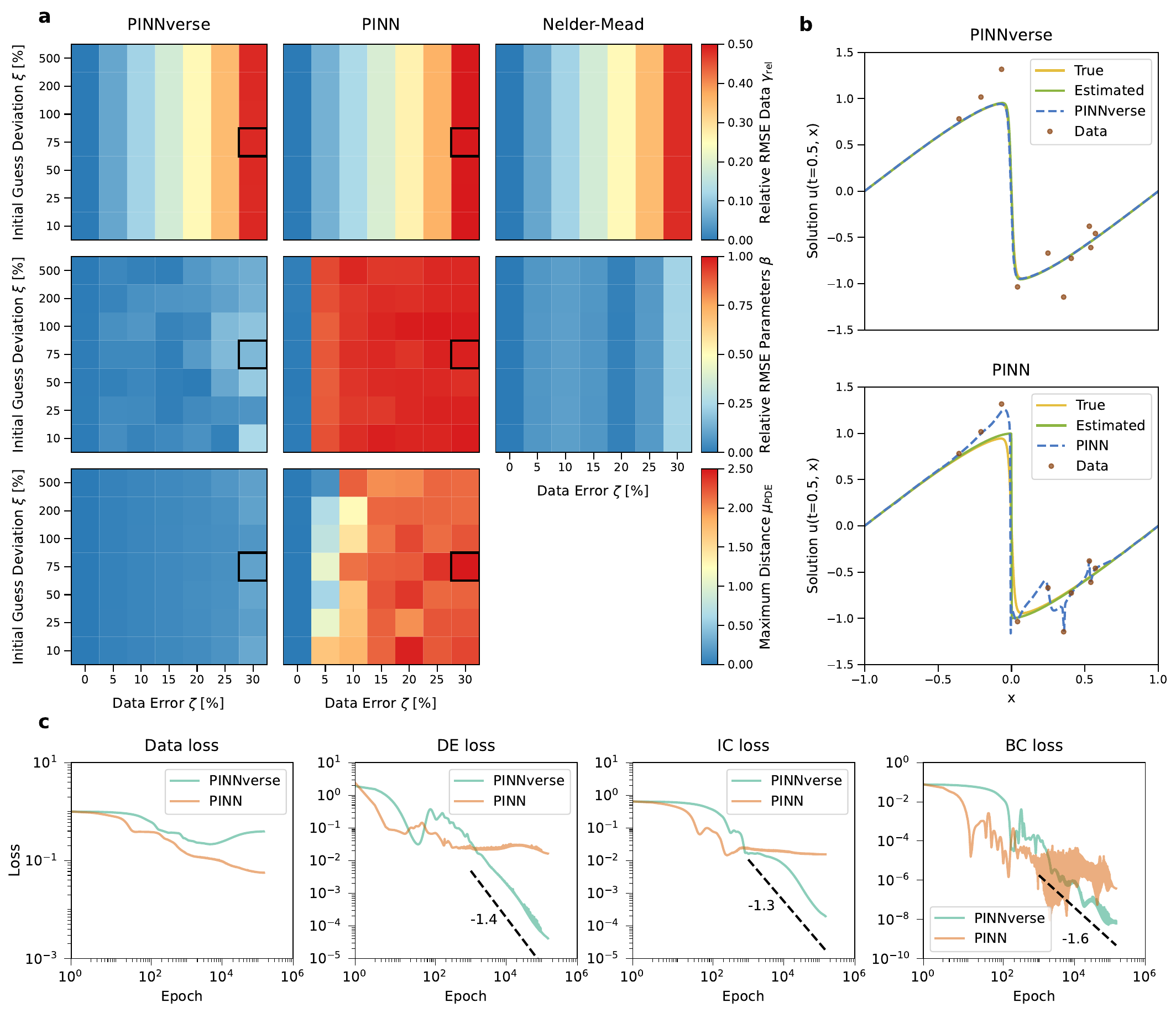}
    \caption{\textbf{Parameter estimation performance in Burgers' PDE model.}
    \textbf{a}, Heatmaps depicting performance metrics across varying noise levels in the data, $\zeta$, and deviations in initial parameter guesses, $\xi$ (Methods). The black square highlights the scenario $\zeta=25\%$, $\xi=75\%$ analyzed in detail in subsequent panels.
    \textbf{b}, Comparison of trajectories for the dependent variable $u(x)$ at time point $t=0.5$, generated using estimated parameters (green curve), true parameters (yellow curve), neural network predictions (blue curve), and the corresponding noisy observational data (brown dots).
    \textbf{c}, Training loss evolution for PINNverse and conventional PINN. Data, differential equation (DE), initial condition (IC) and boundary condition (BC) losses are depicted. For PINNverse, a power law was fitted after 1000 epochs to the DE, IC and BC losses (shifted dashed lines) with indicated exponents.}
    \label{fig:burgers}
\end{figure*}

As a final demanding PDE benchmark, we test PINNverse on Burgers' equation \cite{burgers1948}, which has been widely employed to study various nonlinear wave phenomena, including water infiltration into soil \cite{basha2002}, nonlinear acoustic shock-wave propagation validated by experimental N-waveforms from spark sources \cite{yuldashev2010}, and shock-wave propagation in the human brain resulting from explosive blasts \cite{mason2018}. Burgers' equation has frequently been featured in recent physics-informed neural network literature \cite{raissi2019physics, yu2021, wu2022, das2024}. We use its viscous form
\begin{align*}
\frac{\partial u}{\partial t} + u\,\frac{\partial u}{\partial x} &= \nu\,\frac{\partial^2 u}{\partial x^2},\\
u(0,x) &= -\sin(\pi x)\\
u(t,x)&=0 \quad \text{at}\quad x \in \{-1,1\}
\end{align*}
where $u(t,x)$ represents a velocity field (or an analogous quantity, such as traffic density), and $\nu$ is the viscosity (or diffusion) coefficient. When $\nu > 0$, the solutions are smooth, balancing nonlinearity and diffusion. In the limit $\nu \to 0$, shock waves can form, making Burgers' equation a prototypical model to study shock formation and related phenomena in fluid dynamics.

To generate our synthetic measurements, we selected a ground-truth parameter $\bm{\eta}^{\text{true}} = \mathrm{\nu} = 0.01$, placing the system firmly within the shock wave regime. Lower and upper parameter limits were set to $\bm{\eta}^{\textrm{lower}} = 0$ and $\bm{\eta}^{\textrm{upper}} = 0.07$, respectively. A total of 14 perturbed data points were recorded at observation times $t=0.2$ and $t=0.4$. To overcome the inherent limitation of standard neural networks in representing high-frequency spatial variations, known as spectral bias \cite{tancik2020}, we applied a Fourier feature mapping to the spatial input coordinate (Methods). 150,000 epochs were used for training.

All three optimization strategies achieve similar RMSEs (Fig.~\ref{fig:burgers}a, top row). However, in terms of parameter estimation accuracy (Fig.~\ref{fig:burgers}a, middle row), PINNverse consistently surpasses the standard PINN across all examined scenarios, achieving, on average, a 33-fold improvement in $\beta$. The PINN delivers reliable parameter estimates exclusively under noise-free conditions; as soon as measurement noise is introduced, the accuracy of parameter estimates drastically declines, underscoring the difficulty associated with parameter estimation in shock wave regimes, where inaccurate data can hinder recovery of system dynamics. 

Although Nelder--Mead does not become trapped in local minima within the tested range of initial guesses, it consistently produces less accurate parameter estimates compared to PINNverse. Quantitatively, PINNverse achieves a two-fold mean improvement in $\beta$ across all evaluated scenarios.

PINNverse accurately captures the shock even under noisy conditions, whereas the PINN approach displays substantial deviations and overfitting (Fig.~\ref{fig:burgers}b). Consistent with all previous benchmarks, PINNverse reduces the physical losses in the Fisher--KPP equation approximately algebraically with increasing training effort: $L_\mathrm{de,ic,bc}\sim\text{epoch}^{-a}$ with $a=1.4348\pm0.0003$, $a=1.2761\pm0.0006$ and $a=1.643\pm0.001$ (s.e.) for DE, IC and BC, respectively (Fig.~\ref{fig:burgers}c).

\section*{Discussion}

In the wake of the recent boom of deep learning and artificial intelligence, Physics-Informed Neural Networks have emerged as promising tools that can learn to reproduce and predict solutions to differential equations \cite{jamili2021parameter, mao2020, cai2022, cai2021, cuomo2022scientific}. However, our investigation reveals fundamental limitations within the standard PINN paradigm when applied to parameter inferrence from noisy observational data subject to physical constraints. The core issue lies in the inherent trade-off built into the conventional PINN training process: a simultaneous pursuit of data fidelity and physics compliance, often mediated by fixed weighting schemes that struggle to navigate complex Pareto fronts.

We have introduced PINNverse, a reformulated training paradigm that treats parameter estimation as a constrained optimization problem. By leveraging the Modified Differential Method of Multipliers, we achieve robust convergence to any point on the Pareto front, including those residing within concave regions typically inaccessible to standard descent-based methods. This represents a critical distinction: PINNverse does not just approximate data; it seeks solutions that simultaneously satisfy both the observational constraints and the governing physical equations---a fundamentally different approach than the weighted-sum strategy employed in conventional PINNs.

This key difference became apparent in the solutions predicted by the neural networks: standard PINNs produced substantially distorted solutions that overfit the noisy data, whereas PINNverse accurately reproduced the underlying dynamics. Similar trends were observed across all models considered. PINNverse’s ability to converge on well-balanced regions of the Pareto front was being particularly pronounced in the Burgers equation, where shock wave formation is a defining characteristic. We observed superlinear convergence in the physical constraints (exponents 1.2--1.6) for PINNverse across all studied systems, whereas regular PINNs did not converge.

PINNverse was less sensitive to initial parameter guesses than Nelder--Mead. Moreover, PINNverse naturally supports parameter bounds by simply adding them to the constraints of the differential equation, initial condition and boundary condition loss. Additional constraints can be employed in the same way, further restricting the space of possible solutions. Remarkably, achieving these benefits requires minimal changes---just a few lines of code---to convert an existing standard PINN implementation to PINNverse, and does not inflict noteworthy additional computational cost.

Our findings echo the established efficacy of MDMM in other areas. Recent applications include refining CMS FastSim particle simulation accuracy \cite{bein2024}, Bayesian Entropy Neural Networks with physics constraints \cite{rathnakumar2024}, autonomous robotic cutting \cite{heiden2021} and the inference/pruning of large pre-trained language models \cite{kumar2021, xia2024}.

Despite its demonstrated strengths, PINNverse, as a neural network-based method, inherently depends on careful selection and tuning of hyperparameters such as network architecture (number of layers, neurons per layer) and learning rate. However, in this study, explicit hyperparameter tuning was not conducted, suggesting an inherent resilience of the method to suboptimal parameter configurations.
 
A further strength of PINNverse, although not explicitly demonstrated here, is its computational consistency across diverse problem domains. Unlike traditional numerical methods which typically possess polynomial time complexity in the used spatio-temporal discretization, the computational performance of PINNverse depends primarily on neural network training time. With PINNverse, inverse problems with differential equations can be solved without requiring a single forward evaluation. This suggests a complexity crossover threshold beyond which our method becomes computationally advantageous compared to conventional numerical solvers, particularly for problems requiring fine discretizations, involving complex geometries, multiphysics interactions, or high-dimensional parameter spaces. This theoretical advantage deserves quantitative investigation in future research on increasingly complex models and datasets.

Our investigation utilized a ``vanilla'' PINN architecture as the foundational element of the PINNverse framework. However, the inherent modularity of this approach facilitates seamless integration with various known enhancements to conventional PINNs, e.g. respecting temporal causality \cite{wang2022}, curriculum-based training \cite{krishnapriyan2021}, including Fourier features in the first layer \cite{tancik2020}, adaptive resampling strategies during training \cite{wu2022, daw2023}, and augmenting the training process with additional equations \cite{yu2021}. We anticipate that systematic exploration of these supplementary techniques could yield further refinements in accuracy, computational efficiency, and overall performance.

\section*{Methods}

\subsection*{Solving the inverse problem with Physics-Informed Neural Networks}

We consider a general differential equation (DE) system represented in a residual form given by
\begin{align*}
\mathcal{F}(\bm{x}, t, \bm{u}, \bm{\eta}, \bm{u}_t, \nabla\bm{u}, ...) = 0, \quad &\bm{x} \in \Omega, \quad t \in [0, T]\\
\mathcal{B}(\bm{u}(\bm{x}, t), \bm{x}, t) = 0, \quad &\bm{x} \in \partial\Omega, \,\, t \in [0, T]\\
\bm{u}(\bm{x}, 0) = \bm{h}(\bm{x}), \quad &\bm{x} \in \Omega
\end{align*}
where $\Omega \subseteq \mathbb{R}^n$ represents the spatial domain with boundary $\partial\Omega$, and $\bm{u}: \Omega \times [0, T] \rightarrow \mathbb{R}^m$ denotes the solution field over the space-time domain. The operator $\mathcal{F}(\cdot)$ is a spatio-temporal differential operator encapsulating the governing physics of the system, which may incorporate multiple parameters $\bm{\eta}\in\mathbb{R}^p$ and various spatial and temporal derivatives of $\bm{u}$.
The boundary conditions are imposed through the spatio-temporal operator $\mathcal{B}(\bm{u}, \bm{x}, t)$, which acts on the solution at the domain boundary $\partial\Omega$.
The initial solution $\bm{h}(\bm{x})$ prescribes the state of the system at time $t=0$ throughout $\Omega$.

The forward problem consists of determining the solution $\bm{u}(\bm{x}, t; \bm{\eta})$, given known parameters $\bm{\eta}$. The inverse problem regards the parameters $\bm{\eta}$ as unknown quantities, requiring inference from observational data at discrete spatio-temporal locations. To formalize this inverse scenario, we assume the availability of a dataset comprising $N_{\text{data}}$ observations:
\begin{equation*}
\{(\bm{x}^{\textrm{data}}_i , t^{\textrm{data}}_i, \bm{u}^{\textrm{data}}_i)\}_{i=1}^{N_{\textrm{data}}},
\end{equation*}
where each datum consists of a coordinate pair $(\bm{x}_i^{\text{data}}, t_i^{\text{data}})$ and the corresponding observed solution $\bm{u}_i^{\text{data}}$. The goal is to approximate the solution using a NN model $\bm{u}^{\bm{\theta}}(\bm{x}, t)$, parameterized by neuronal weights and biases collectively denoted by $\bm{\theta}$. The optimal parameters are determined by minimizing the discrepancy between the network predictions and observed data in some chosen metric:
\begin{equation*}
\bm{\theta}^* = \argmin_{\bm{\theta}} L_{\text{data}}(\bm{\theta}).
\end{equation*}
We seek the minimum in a least-squares sense here. The data loss function $L_{\textrm{data}}(\bm{\theta})$ can be formulated as either an absolute loss function
\begin{equation*}
L_{\textrm{data}}(\bm{\theta}) = \sqrt{\frac{1}{N_{\textrm{data}}} \sum_{i=1}^{N_{\textrm{data}}} \left(\bm{u}^{\bm{\theta}} (\bm{x_i},t_i) - 
\bm{u}^{\textrm{data}}_i\right)^2}
\end{equation*}
or, unless the data approaches zero, as a relative loss function
\begin{equation*}
L_{\textrm{data}}(\bm{\theta}) = \sqrt{\frac{1}{N_{\textrm{data}}} \sum_{i=1}^{N_{\textrm{data}}} \left(\frac{\bm{u}^{\bm{\theta}} (\bm{x_i},t_i) - 
\bm{u}^{\textrm{data}}_i}{\bm{u}^{\textrm{data}}_i}\right)^2}
\end{equation*}
where the vector division is taken element-wise.

To impose conformity with the physical laws described by the DE, a residual loss is introduced as
\begin{equation*}
L_{\textrm{de}}(\bm{\theta}, \bm{\eta}) = \frac{1}{N_c} \sum_{i=1}^{N_c} \mathcal{F}(\bm{x_i}^\textrm{de}, t_i^\textrm{de}, \bm{u}^{\theta}(\bm{x_i}^\textrm{de}, t_i^\textrm{de}), \bm{\eta}, \ldots)^2,
\end{equation*}
where $\{{\bm{x}_i^\textrm{de}, t_i^\textrm{de}}\}_{i=1}^{N_c}$ are collocation points sampled within $\Omega \times (0,T)$. 

Additionally, losses for initial and boundary conditions are defined by
\begin{align*}
L_{\textrm{bc}}(\bm{\theta}) &= \frac{1}{N_{\textrm{bc}}} \sum_{i=1}^{N_{\textrm{bc}}} \mathcal{B}\left(\bm{u}^{\bm{\theta}}(\bm{x}_i^\mathrm{bc}, t_i^\mathrm{bc}),\bm{x}_i^\mathrm{bc}, t_i^\mathrm{bc}\right) ^2\\
L_{\textrm{ic}}(\bm{\theta}) &= \frac{1}{N_{\textrm{ic}}} \sum_{i=1}^{N_{\textrm{ic}}} \left(\bm{u}^{\bm{\theta}}(\bm{x}_i^\textrm{ic}, 0) - \bm{h}(\bm{x}_i^\textrm{ic}) \right)^2 
\end{align*}
where $\{\bm{x}_i^\mathrm{bc}, t_i^\mathrm{bc}\}_{i=1}^{N_{\textrm{bc}}}$ are boundary condition points in $\partial\Omega \times [0, T]$ and
$\{\bm{x}_i^\textrm{ic},0\}_{i=1}^{N_{\textrm{ic}}}$ are initial conditional points in $\Omega \times \{t = 0\}$.

Traditionally, a composite total loss function is then formulated as \cite{raissi2019physics}
\begin{align*}
L_{\textrm{pinn}}(\bm{\theta}, \bm{\eta}) &= \omega_{\textrm{data}} L_{\textrm{data}}(\bm{\theta}) + \omega_{\textrm{de}} L_{\textrm{de}}(\bm{\theta}, \bm{\eta})\\ 
&+\omega_{\textrm{ic}} L_{\textrm{ic}}(\bm{\theta}) + \omega_{\textrm{bc}} L_{\textrm{bc}}(\bm{\theta})
\end{align*}
where $\omega_{\textrm{data}}$, $\omega_{\textrm{de}}$, $\omega_{\textrm{ic}}$ and $\omega_{\textrm{bc}}$ are weights that balance the partial losses. Following common practice \cite{raissi2019physics}, all weights were set to one here.

PINNs leverage automatic differentiation to compute derivatives of the output variables $\bm{u}$ with respect to $\bm{x}$ and $t$, enabling evaluation of the differential operators $\mathcal{F}(\cdot)$ and boundary operators $\mathcal{B}(\cdot)$. The parameter update using gradient descent is performed as
\begin{align*}
\bm{\theta}^{(k+1)} &= \bm{\theta}^{(k)} - \alpha \nabla_{\bm{\theta}} L_{\textrm{pinn}} (\bm{\theta}^{(k)}, \bm{\eta}^{(k)})\\
\bm{\eta}^{(k+1)} &= \bm{\eta}^{(k)} - \alpha \nabla_{\bm{\eta}} L_{\textrm{pinn}} (\bm{\theta}^{(k)}, \bm{\eta}^{(k)})
\end{align*}
where $\alpha>0$ is the learning rate and $k$ the iteration index.

\subsection*{Pareto optimality}

With noisy experimental data, the composite loss function encompasses multiple competing objectives. This represents a multi-objective optimization problem
\begin{equation*}
\min_{\Psi} \bm{L}(\bm{\Psi}).
\end{equation*}
where we seek to simultaneously optimize all components of a total loss vector
\begin{align*}
\bm{L}(\bm{\Psi}) &= \bm{L}(\bm{\theta}, \bm{\eta})\\
&= (L_{\textrm{data}}(\bm{\theta}), L_{\textrm{de}}(\bm{\theta}, \bm{\eta}), L_{\textrm{ic}}(\bm{\theta}), L_{\textrm{bc}}(\bm{\theta}))^\mathrm{T}.
\end{align*}
Finding a single parameter vector $\bm{\Psi}$ that simultaneously minimizes all loss components is generally infeasible. To formalize this challenge, we adopt the concept of Pareto optimality. A parameter vector $\bm{\Psi}$ is considered (globally) Pareto optimal, if no other parameter vector $\bm{\Psi}$ exists that achieves non-increasing values across all loss functions while strictly improving at least one loss component. The collection of all candidate solutions constitutes the feasible region.

The subset of optimal objective function values represents the Pareto front \cite{pareto1971} (Fig.~\ref{fig:PINNverse}, solid black curve), which manifests in two fundamental geometric configurations: convex and concave. A convex Pareto front is distinguished by the property that for any two points $\bm{a}$ and $\bm{b}$ on the front and any scalar $\kappa \in [0,1]$, there exists a point $\bm{c}$ on the front such that $\kappa||\bm{a}|| + (1 - \kappa)||\bm{b}|| \geq ||\bm{c}||$. Conversely, a concave Pareto front satisfies the inequality $\kappa||\bm{a}|| + (1 - \kappa)||\bm{b}|| \leq ||\bm{c}||$.

The performance of gradient-based optimization methods is intrinsically linked to the geometry of the Pareto front. Specifically, when minimizing linearly weighted objectives, gradient descent converges exclusively to solutions located on the convex regions of the Pareto front \cite{das1997}. Consequently, regardless of the positive weighting parameters selected, points within non-convex segments of the front cannot be attained, as they do not correspond to minima of any weighted sum objective function. In contrast, for purely convex Pareto fronts, gradient-based optimization can theoretically converge to any desired point along the curve through appropriate adjustment of the weighting parameters. However, precisely controlling the final solution point along the front via weight selection is often non-trivial, as the mapping between weights and Pareto points is highly sensitive to the front's local curvature \cite{das1997}.

In practical applications with neural networks, Pareto fronts typically have mixed shapes with both convex and concave regions, making the tuning of PINNs notoriously difficult \cite{wong2025}.

\subsection*{Inverse problem as constraint optimization}

To address the limitations of standard gradient-based methods on complex Pareto fronts, we reformulate the PINN training process as a constrained optimization problem. Rather than treating all loss terms equally, we designate the data-fitting term as the primary objective while transforming the physics-based terms into constraints:
\begin{align*}
    \underset{\bm{\theta}}{\text{minimize}} \quad &L_{\textrm{data}}(\bm{\theta}) \\
    \text{subject to} \quad &L_i(\bm{\theta}, \bm{\eta}) = 0, \quad i \in \mathcal{I}_\mathrm{e}=\{\mathrm{de}, \mathrm{ic}, \mathrm{bc}\} \\
    \qquad \qquad \eta^{\textrm{lower}}_j \leq &\; \eta_j \leq \eta^{\textrm{upper}}_j, \quad j \in \mathcal{I}_\mathrm{b} = \{1,\ldots ,p\}
\end{align*}
The parameters $\eta_j\in$ represent differential equation parameters constrained within physically plausible bounds $[\eta^{\textrm{lower}}_j, \eta^{\textrm{upper}}_j]$. These bounds ensure that the solution remains physically meaningful and prevent the neural network from exploring invalid regions.

To handle the bound constraints efficiently, we introduce an infeasibility function
\begin{equation*}
V_j(\eta_j(\bm{\theta}))=\max(\eta^{\textrm{lower}}_j,\min(\eta_j(\bm{\theta}) ,\eta^{\textrm{upper}}_j)) - \eta_j(\bm{\theta})
\end{equation*}
that measures constraint violations. This allows us to express the Lagrangian as
\begin{align*}  
\mathcal{L}(\bm{\theta}, \bm{\eta}, \bm{\lambda}, \bm{\chi}) &= L_{\textrm{data}}(\bm{\theta}) + \sum_{i \in \mathcal{I}_{e}}\lambda_i L_i(\bm{\theta}, \bm{\eta}) \\
&+ \sum_{j \in \mathcal{I}_{b}} \chi_j V_j(\eta_j),
\end{align*}
where $\lambda_i$ and $\chi_j$ represent the Lagrange multipliers of equality and parameter bound constraints.
The optimal set of neural network parameters is then obtained through a min-max formulation:
\begin{equation*}
(\bm{\theta}, \bm{\eta})^* = \argmin_{\bm{\theta}, \bm{\eta}}\,\left(\max_{\bm{\lambda} \geq 0, \bm{\chi} \geq 0} \mathcal{L}( \bm{\theta}, \bm{\eta}, \bm{\lambda}, \bm{\chi})\right).
\label{eq:MiniLagrange}
\end{equation*}
The target solution for the Lagrangian min-max formulation is inherently a saddle point \cite{boyd2004}. However, such points are generally not attractors for standard gradient-based optimizers \cite{platt1987}. 

\subsection*{Optimization approach of PINNverse}

To ultimately overcome these limitations, we employ the Modified Differential Method of Multipliers (MDMM) \cite{platt1987}. To the best of our knowledge, this represents the first application of the MDMM in the context of PINNs. MDMM is an optimization algorithm derived from the augmented Lagrangian formulation, also known as the Method of Multipliers. This formulation introduces quadratic penalty terms alongside the standard Lagrange multiplier terms to improve convergence properties. For the PINNverse problem, the augmented Lagrangian is defined as
\begin{align*}
   \mathcal{L}_\mathrm{A}(\bm{\theta}, \bm{\eta},\bm{\lambda}, \bm{\chi}, \bm{c}) &=
    L_{\textrm{data}}(\bm{\theta})\\
    &+ \sum_{i \in \mathcal{I}_{\text{e}}} \left( \lambda_i L_i(\bm{\theta},\bm{\eta}) + \frac{c_i}{2} L_i^2(\bm{\theta},\bm{\eta})\right) \\
   & + \sum_{j \in \mathcal{I}_{\text{b}}} \left( \chi_j V_j(\eta_j) + \frac{d_j}{2} V_j^2(\eta_j) \right),
\end{align*}
where $c_i>0$ and $d_j>0$ are the penalty coefficients for the constraints. Larger values enforce constraints more strictly. In this study, all penalty parameters were set to unity ($c_i=d_j=1$).

A key distinction of MDMM from standard sequential augmented Lagrangian methods lies in its update dynamics. MDMM proposes simultaneous updates for both the primal variables ($\bm{\theta}$, $\bm{\eta}$) and the Lagrange multipliers ($\bm{\lambda}$, $\bm{\chi}$). For a gradient descent update this reads
\begin{align*}
   \bm{\theta}^{(k+1)} &= \bm{\theta}^{(k)} - \alpha \nabla_{\bm{\theta}} \mathcal{L}_\mathrm{A}(\bm{\theta}^{(k)}, \bm{\eta}^{(k)}, \bm{\lambda}^{(k)}, \bm{\chi}^{(k)})\\
  \bm{\eta}^{(k+1)} &= \bm{\eta}^{(k)} - \alpha \nabla_{\bm{\eta}} \mathcal{L}_\mathrm{A}(\bm{\theta}^{(k)}, \bm{\eta}^{(k)}, \bm{\lambda}^{(k)}, \bm{\chi}^{(k)}).
\end{align*}
Here $\alpha>0$ represents the learning rate that controls the step size during each iteration of the gradient descent.
Crucially, in MDMM the Lagrange multipliers are updated via gradient ascent:
\begin{align*}
   \lambda_i^{(k+1)} &= \lambda_i^{(k)} + \alpha\, L_i(\bm{\theta}^{(k)}, \bm{\eta}^{(k)}) , \quad i \in \mathcal{I}_{\text{e}}\\
   \chi^{(k+1)}_j &= \chi^{(k)}_j + \alpha\, V_j(\eta_j^{(k)}) , \quad j \in \mathcal{I}_{\text{b}}
\end{align*}

The inclusion of the quadratic penalty term, governed by $c_i$, $d_j$, is essential. As established in optimization theory \cite{nocedal2006, platt1987}, for sufficiently large penalty parameters, the Hessian of the augmented Lagrangian with respect to the primal variables ($\nabla_{\bm{\theta}, \bm{\xi}}^2 \mathcal{L}_\mathrm{A}$) becomes positive definite in the subspace tangent to the constraints near a constrained minimum satisfying standard second-order sufficiency conditions. This induces local convexity and transforms the constrained minimum into an attractor for the dynamics, mitigating the saddle-point issues associated with the standard Lagrangian that hinder simple gradient descent \cite{platt1987}. Note that we still need gradient ascent for the Lagrange multipliers, since the convexity only holds for the primal variables.

Consequently, MDMM offers robust convergence towards a constrained minimum for sufficiently large $c_i$ and $d_j$, suitable learning rate $\alpha$, and initialization within the basin of attraction. Notably, this minimum can be any point on the Pareto front, even in the non-convex region.

In the theoretical derivation presented above, gradient updates were illustrated using stochastic gradient descent for simplicity. However, any gradient-based optimization algorithm is compatible with the MDMM framework. Motivated by this flexibility, we adopt the recently proposed Adan optimizer \cite{xie2024}, an adaptive optimization technique grounded in Nesterov Momentum Estimation, specifically tailored for rapid and stable convergence in non-convex optimization landscapes.

\subsection*{Training}

For all presented results, both the standard PINN and PINNverse were trained using neural networks comprising two hidden layers, each consisting of 20 neurons, with hyperbolic tangent activation functions. We employed a learning rate scheduler characterized by an initial linear decay from $\alpha = 10^{-2}$ down to $10^{-4}$ until reaching the last 30,000 epochs, after which the learning rate was kept constant at $\alpha = 10^{-4}$. For discretization, $N_{\mathrm{de}} = 16,384$ collocation points were uniformly distributed across the interior of the temporal or spatio-temporal domains using a Sobol sequence \cite{sobol1967}. An exception was the FitzHugh--Nagumo model, for which we used $N_{\mathrm{de}} = 10,000$. In the two PDEs, additional collocation points, specifically $N_{\mathrm{ic}} = N_{\mathrm{bc}} = 1,024$, were allocated to enforce the initial and boundary conditions, respectively.

For Burgers' equation, Fourier features were used in the training. This technique transforms the coordinate into a higher-dimensional feature space using sinusoidal basis functions. Ten such basis functions corresponding to distinct frequencies were employed in our network. This augmented spatial representation, concatenated with the temporal coordinate, served as the network input, thereby enhancing its capability to resolve the sharp gradients of shock wave dynamics. 

\subsection*{Evaluation of accuracy}

We evaluated method performance under realistic conditions by introducing heteroscedastic Gaussian noise to the data,
\begin{equation*}
\hat{y} \sim \mathcal{N}\left(y, \zeta y\right)
\end{equation*}
with noise levels $\zeta$ up to 30\%. Additionally, we used substantially perturbed initial guesses for parameter initialization,
\begin{equation*}
\bm{\eta}^{\textrm{start}} = (1+\xi) \bm{\eta}^{\textrm{true}}
\end{equation*}
with relative deviations $\xi$ up to 500\%.

To quantitatively evaluate solution accuracy, we define the maximum distance metric $\mu$. For ODE problems, this metric is formulated as
\begin{equation*}
\mu_{\text{ODE}} = \max_{\substack{t \in [0,T]\\i \in \{1,\ldots,m\}}} \left|u^{\text{NN}}_i(t; \bm{\theta}, \bm{\eta}) - u^{\text{true}}_i(t; \bm{\eta}^{\text{true}})\right|
\end{equation*}
where $u^{\text{NN}}_i(t; \bm{\theta})$ the neural network prediction with parameters $\bm{\theta}$ for the $i$-th solution component at time $t$, and $u^{\text{true}}_i(t; \bm{\eta}^{\text{true}})$ denotes the corresponding true solution with parameters $\bm{\eta}^{\text{true}}$ obtained via high-precision numerical methods. For PDE problems, we extend this metric to incorporate spatial dimensions:
\begin{equation*}
\mu_{\text{PDE}} = \max_{\substack{t \in \mathcal{T}\\\bm{x} \in \Omega\\i \in \{1,\ldots,m\}}} \left|u^{\text{NN}}_i(t,\bm{x}; \bm{\theta}, \bm{\eta}) - u^{\text{true}}_i(t,\bm{x}; \bm{\eta}^{\text{true}})\right|
\end{equation*}
where $\Omega$ is the spatial domain and $\mathcal{T}$ represents the discrete set of measured time points. Note that for the PDE case we only consider the discrete time points where we have measurements, not the whole time domain. A well-trained model that adheres to the underlying physics should yield $\mu$ values approaching zero.

To assess the parameter estimation performance of the three techniques, we computed the relative root mean squared error between the true parameters and the estimated parameters:
\begin{equation*}
\beta = \sqrt{\frac{1}{p}\sum_{j=1}^{p}\left(\frac{\eta_j^{\textrm{true}}-\eta_j^{\textrm{est}}}{\eta_j^{\textrm{true}}}\right)^2}
\end{equation*}
where $p$ denotes the total number of parameters in the differential equation.

Additionally, we evaluated the model performance by comparing the noisy observed data with the predictions obtained by solving the differential equations using the estimated parameters in absolute and relative terms:
\begin{align*}
    \gamma_{\textrm{abs}} &= \sqrt{\frac{1}{N_{\textrm{data}}}\sum_{j=1}^{N_{\textrm{data}}}\big(\hat{\bm{y}}_j-\bm{u}^{\textrm{pred}}(t_j, \bm{x}_j; \bm{\eta}^{\textrm{est}})\big)^2}\\
    \gamma_{\textrm{rel}} &= \sqrt{\frac{1}{N_{\textrm{data}}}\sum_{j=1}^{N_{\textrm{data}}}\left(\frac{\hat{\bm{y}}_j-\bm{u}^{\textrm{pred}}(t_j, \bm{x}_j; \bm{\eta}^{\textrm{est}})}{\hat{\bm{y}}_j}\right)^2}
\end{align*}
where $N_{\textrm{data}}$ represents the total number of data points, $\hat{\bm{y}}_j$ denotes the $j$-th noisy measurement vector, and $\bm{u}^{\textrm{pred}}(t_j, \bm{x}_j; \bm{\eta}^{\textrm{est}})$ is the predicted vector at the corresponding space-time point using the estimated parameters $\bm{\eta}^{\textrm{est}}$.

\bmhead{Data availability}
No new data was generated for this study.

\bmhead{Code availability}
The complete PyTorch code, which which all our results can be reproduced, is freely available as a git repository at \url{https://git.bsse.ethz.ch/iber/Publications/2025_almanstoetter_pinnverse}.

\bmhead{Acknowledgements}
None.

\bmhead{Competing interests}
The authors declare no competing interests.

\bibliography{references}

\end{document}